\documentclass[letterpaper, 10 pt, conference]{ieeeconf}  

\IEEEoverridecommandlockouts                              

\usepackage{graphics} 
\usepackage{epsfig} 
\usepackage{caption}
\usepackage{subfigure}
\usepackage{url}
\usepackage{multirow}
\usepackage{balance}
\usepackage{amsmath} 
\usepackage{amssymb}  
\begin{document}
\title{\LARGE \bf
	Grasp State Assessment of Deformable Objects Using Visual-Tactile Fusion Perception
}

\author{Shaowei Cui$^{1, 2}$, Rui Wang$^{1}$, Junhang Wei$^{1, 2}$, Fanrong Li $^{1, 2}$ and Shuo Wang$^{1, 2, 3}$
	\thanks{This work was supported in part by by the National Natural Science Foundation of China under Grant U1713222, Grant 61773378. corresponding author: {\tt\small shuo.wang@ia.ac.cn}}
	\thanks{$^{1}$ S. Cui, R. Wang, J. Wei, and S. Wang are with the State Key Laboratory of Management and Control for Complex Systems, Institute of Automation, Chinese Academy of Sciences, Beijing 100190, China.}%
	\thanks{$^{2}$ S. Cui, J. Wei, and F. Li are with School of Further Technology, Universty of Chinese Academy of Sciences, Beijing 100190, China.}
	\thanks{$^{3}$ S. Wang is also with Center for Excellence in Brain Science and Intelligence Technology Chinese Academy of Sciences, Shanghai 200031, China.}
	%
}

	\maketitle
	\thispagestyle{empty}
	\pagestyle{empty}

\begin{abstract}
Humans can quickly determine the force required to grasp a deformable object to prevent its sliding or excessive deformation through vision and touch, which is still a challenging task for robots. To address this issue, we propose a novel 3D convolution-based visual-tactile fusion deep neural network (C3D-VTFN) to evaluate the grasp state of various deformable objects in this paper. Specifically, we divide the grasp states of deformable objects into three categories of sliding, appropriate and excessive. Also, a dataset for training and testing the proposed network is built by extensive grasping and lifting experiments with different widths and forces on 16 various deformable objects with a robotic arm equipped with a wrist camera and a tactile sensor. As a result, a classification accuracy as high as $99.97\%$ is achieved. Furthermore, some delicate grasp experiments based on the proposed network are implemented in this paper. The experimental results demonstrate that the C3D-VTFN is accurate and efficient enough for grasp state assessment, which can be widely applied to automatic force control, adaptive grasping, and other visual-tactile spatiotemporal sequence learning problems.
\end{abstract}

\section{Introduction}

Robotic grasp capability is receiving increasing attention due to increased demand for various dexterity grasping and manipulation of service and industrial robots \cite{c1}, \cite{c28}. To improve the general grasp ability of the robots, accurate and efficient grasp state assessment is a relatively critical part. Traditional grasp quality assessment focuses on whether a grasp process is stable and whether slippage has occurred. Many scholars have already researched in the grasp stability assessment \cite{c2}, \cite{c3} and slip detection/prediction \cite{c4}, \cite{c5}. 

\begin{figure}[t!]
	\centering
	\includegraphics[width=8.3cm]{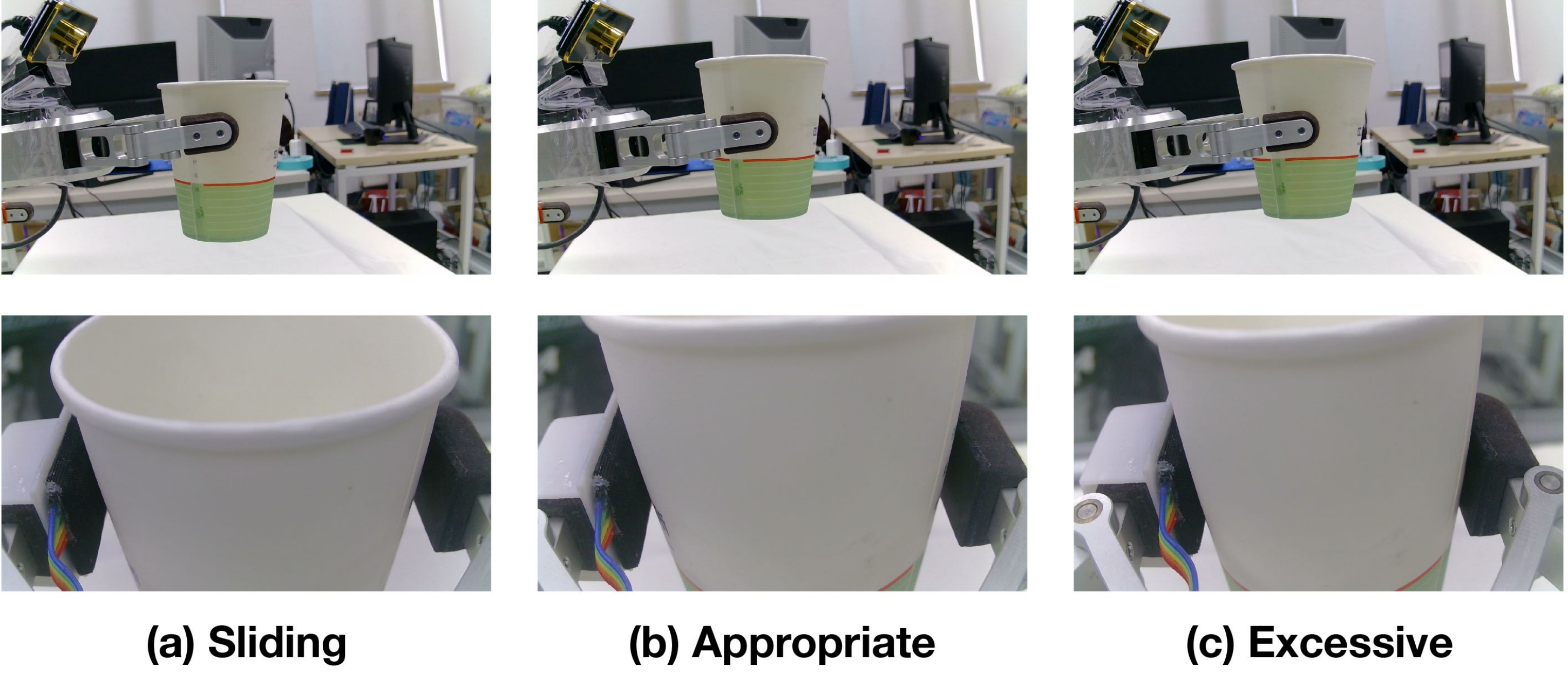}
	\caption{(a): The sliding grasp state. (b) The appropriate grasp state. (c) The excessive grasp state. The top row images are captured by a side-mounted camera and the bottom by a wrist camera.}
	\label{grasp_states}
\end{figure}

Nevertheless, for a deformable or fragile object, it is not enough to only detect whether it slides during a grasp process. For example, for such a task of grasping paper cups, if the gripping force is set too large, although the paper cup can be prevented from slipping, the excessive gripping force may cause the paper cup to undergo a large deformation, thereby causing irreversible damage. To this end, a more comprehensive approach to assess the grasp state of deformable objects needs to be studied. In this paper, we define the grasp state assessment task for deformable objects as a tri-classification problem with sliding, appropriate, and excessive labels. These three grasp states are used to describe the grasp state of various deformable objects, as shown in Fig. \ref{grasp_states}.

Vision and tactile sensing are two of the primary sensing modalities to perceive the ambient world for humans \cite{c6}. Vision provides the appearance, shape, and other visible features of objects, while touch provides more accurate texture, roughness, contact strength, and other invisible details \cite{cui}. For such a grasp state assessment task, humans are capable of intuitively performing the evaluation process. Someone picking up a random object can automatically determine if the grasp is appropriate \cite{c7}. This information benefits from both tactile and visual feedback. The same is true for robots, and  this paper focuses on how to endow robots the ability to evaluate the grasp state of deformable objects using visual-tactile fusion perception.

\begin{figure}[thpb]
	\centering
	\includegraphics[width=6.5cm]{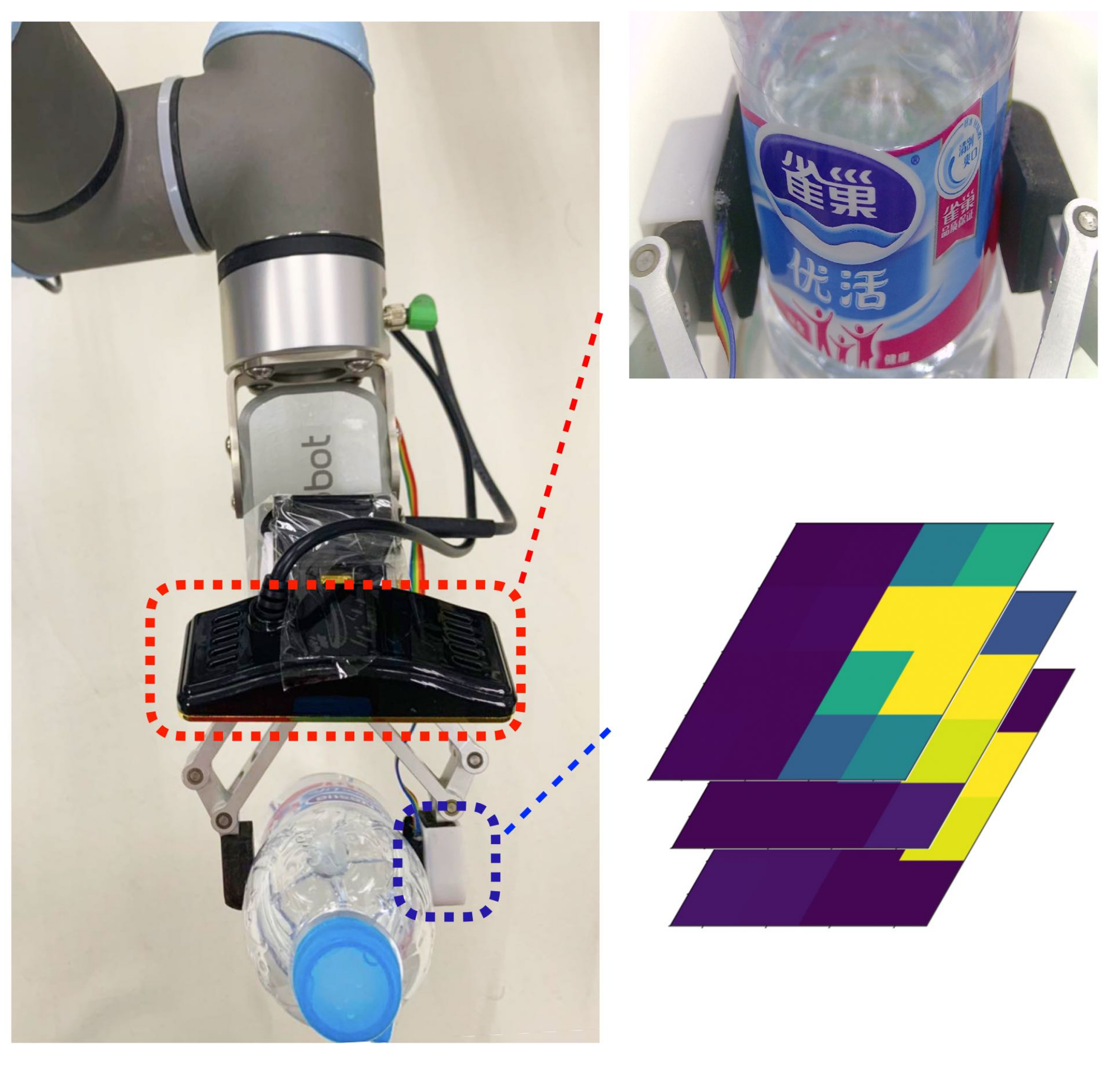}
	\caption{Left: The experiment setup: a UR3 robot arm with an OnRobot RG2 gripper. One finger of the gripper is equipped with a XELA Tactile sensor \cite{c8}. A 1080P USB camera is mounted on the top of the gripper. Upper right: The photo taken by the wrist camera. Bottom right: Three-axis force distribution map from the tactile sensor.}
	\label{robot_setup}
\end{figure}

The difficult primary issue involved in such a bimodal fusion perception task is how to learn effective fusion spatiotemporal features from two heterogeneous modal spatiotemporal sequences \cite{c4}. In this paper, we propose a novel 3D convolution-based visual-tactile fusion deep neural network (C3D-VTFN) to evaluate the grasp state, mimicking the strategy adopted by humans.  
Furthermore, we perform extensive grasping and lifting experiments with different grasp settings to train and test the neural work on our humanoid robot platform. The visual and tactile sequences are taken from a wrist camera fixed above a gripper and a XELA \cite{c8} tactile sensor, respectively. The experimental setup is shown in Fig. \ref{robot_setup}. Finally, some comparative experiments of C3D-VTFN model with different inputs and two real-time grasp state correction experiments based on proposed model are implemented to verify the effectiveness of the proposed network further.

This paper is organized as follows: in Section II, the related work of grasp state assessment and visual-tactile fusion perception are explained. In Section III, the problem statement, detailed architecture of C3D-VTFN, and training specifications are described. In Section IV, the experimental results and discussions are provided. Finally, the contributions of this paper and future work are discussed in Section V.

\section{Related work}

\subsection{Grasp state assessment}
Grasp state assessment is critical for robots to achieve high-quality grasping and manipulation tasks. In the past decades, most studies have focused on the stability in the grasp process. Yasemin Bekiroglu \emph{et al.} \cite{c9} studied the problem of learning grasp stability in robotic object grasping based on tactile measurement and Hidden Markov Models (HMMs). \cite{c10} proposed an integrating grasp planning with an online stability assessment based on tactile sensing. They also presented a probabilistic framework for grasp modeling and stability assessment \cite{c11}. Yevgeb Chebotar \emph{et al.} introduced a framework for learning re-grasping behaviors based on tactile data. They presented a grasp stability predictor that used spatio-temporal tactile features \cite{c12}. 

Moreover, a novel method to incorporate exteroception and proprioception into grasp stability assessment was proposed by \cite{c2}. A convolutional neural network (CNN) was used to extract features and fusion of different modality information. More recently, A new method to predict grasp stability using a non-matrix tactile sensor was proposed by \cite{c13}. Filipe Veiga \emph{et al.} proposed a grip stabilization approach for novel objects based on slip prediction \cite{c5}. Besides, Graph convolutional network method was also studied to predict grasp stability with tactile sensors\cite{c14}. 

Most of the above studies have focused on stability assessment during the grasp process while ignoring dexterity. However, excessive-force gripping can achieve stable grasp of deformable objects but may also cause irreversible damage. Therefore, we conduct a more comprehensive grasp state assessment of daily life deformable objects. By adding the grasp state of excessive detection, a more comprehensive grasp state assessment framework is proposed for more sophisticated and complex robotic grasping and manipulation tasks.

\subsection{Visual-tactile fusion perception}
Vision and tactile sensing are two primary important modalities in robotics perception. In the past decades, it is still challenging to combine vision and touch modalities to facilitate robot manipulations due to their different sensing principles and data structures. However, these limitations have recently improved due to the increasing measurement accuracy of tactile sensors and advances in fusion algorithms with deep neural networks. The combination of visual and tactile perception plays an increasingly important role in the robotic community \cite{c15}. Visual-tactile fusion perception has long been used for a variety of tasks, such as surface classification \cite{c16}, object recognition \cite{c17}, object 3D shape perception \cite{c18}, etc.


In the field of grasping and manipulation, R Calandra \emph{et al.} investigated the question of whether touch sensing aids in predicting grasp outcomes within a multimodal sensing framework that combines vision and touch \cite{c15}. The experimental results indicated that incorporating tactile readings substantially improve grasp performance. Furthermore, an end-to-end action-conditional model that learns re-grasping policies from rowed visual-tactile data was proposed in \cite{c19}. The re-grasping strategy using combined visual and tactile sensing had greatly improved the success of grasping. Michelle A. Lee \emph{et al.} used self-supervision to learn a compact and multimodal representation of RGBs, depth, force-torque, and proprioception for different contact-rich manipulation \cite{c20}.

 Nevertheless, these studies only use tactile and visual images at a specific moment as input and do not use time-domain information of the two modalities. Spatiotemporal features of visual and tactile are extracted by Convolutional Neural Network (CNN)+Recurrent Neural Network (RNN) architecture for slip detection \cite{c4}. However, they only detect slip, and the premise of this study is that the reading frequency of the tactile and visual data is consistent, but most tactile sensors read more quickly than the cameras. Hence, we present a novel C3D grounded framework to tackle the visual-tactile fusion perception problem.

\begin{figure*}[htbp]
	\centering
	\includegraphics[width=12cm]{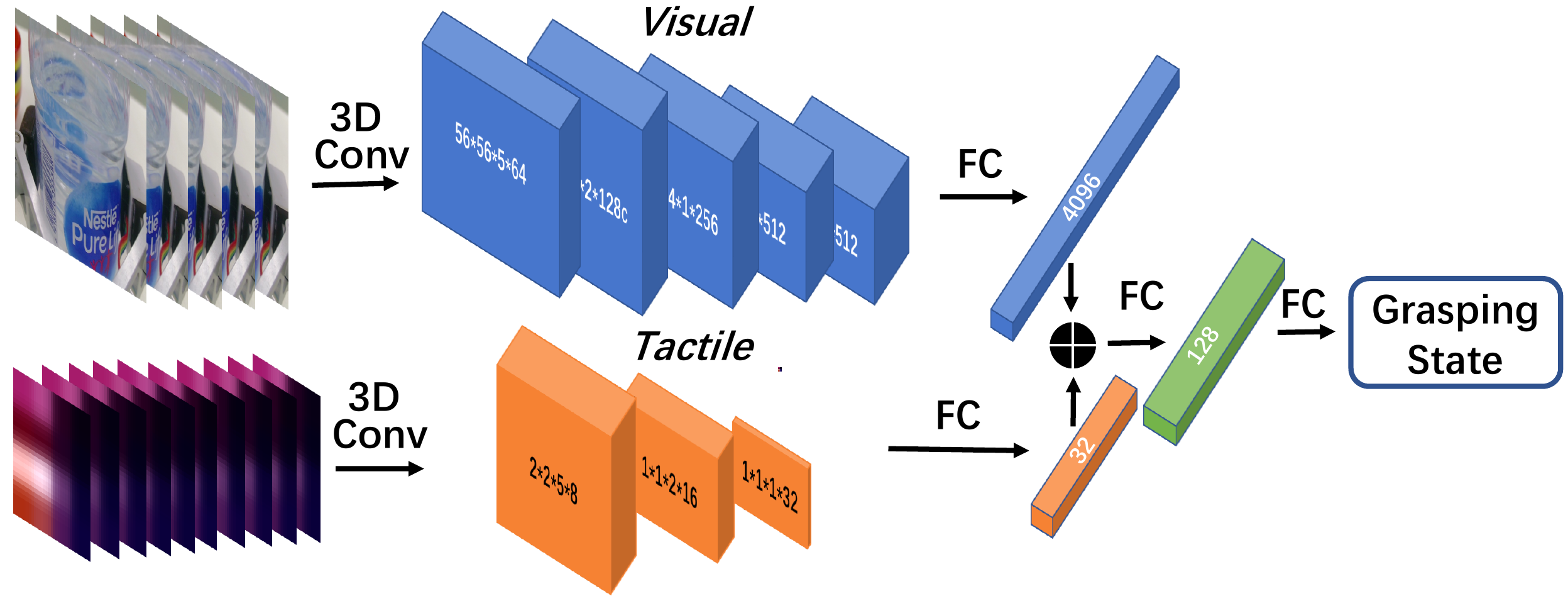}
	\caption{The diagram of C3D-VTFN model. Blue and orange blocks denote visual and tactile 3D convolutional layers, respectively. The cuboids with different colors represent FC layers. }
	\label{modelarch}
\end{figure*}
\section{Proposed method}\label{vtfa-rnn}
\subsection{Problem statement}
Our goal is to obtain the current grasp state by visual-tactile fusion perception. Given the visual ($X_{v1}$, $X_{v2}$,...,$X_{vm}$) and tactile ($X_{t1}$, $X_{t2}$,...,$X_{tn}$)\footnote{$m$ and $n$ are the lengths of the two sequences, respectively.} sequences, we first extract visual ($\mathit{F}_v$) and tactile ($\mathit{F}_t$) features by visual ($E_v$) and tactile ($E_t$) encoder functions and then construct a fusion feature ($\mathit{F}_{v,t}$) based on them. Finally, $\mathit{F}_{v,t}$ is fed into a classification function $\mathbb{F}_{c}$ to predict the current grasp state $y$. This problem is formulated as
\begin{equation}
\mathit{F}_{v,t}=E_v(X_{v1},X_{v2},..,X_{vm})\oplus E_t(X_{t1}, X_{t2},..,X_{tn})
\end{equation}
\begin{equation}
y=\mathbb{F}_{c}(\mathit{F}_{v,t})\; \;y\in\; 0,\,1\, 2
\end{equation}
Which $0$, $1$, and $2$ refer to the sliding, appropriate, and excessive grasp states, respectively. Hence, the grasp state assessment task is defined as a tri-classification problem. 

To address the above problem, we propose a novel 3D Convolution-based visual-tactile fusion network (C3D-VTFN) in this paper, where $E_v$ and $E_t$ are implemented by 3D convolutional neural networks with parameters $\theta_{v}$ and $\theta_{t}$,  and $\mathbb{F}_{c}$ is constructed by Fully-Connection (FC) layers with parameters $\theta_{c}$.
\subsection{Model description}

The overall architecture of C3D-VTFN model is shown in Fig. \ref{modelarch}. The proposed model consists of three components including visual feature extraction module ($E_v$), tactile feature extraction module ($E_t$), and classification module ($\mathbb{F}_{c}$). Given the current visual and tactile sequence, the output of C3D-VTFN is the current grasp state category. Firstly, the visual and tactile features of each spatiotemporal sequence are extracted by C3D networks. Note that tactile modal input is also treated as a small image because of its matrix distribution. Finally, the visual and tactile features are then combined using FC layers to generate a classification result. 

\begin{table}[htbp]
	\caption{Detailed network parameters of C3D-VTFN.}
	\label{c3d-vtfn-para}
	\begin{center}
		\begin{tabular}{|c|c|c|}
			\hline
			 &\textbf{Visual layers}&\textbf{Output size}\\
			\hline
			3d-conv$_1$&3$\times$3$\times$3$\times$64, padding(1,1,1), relu&112$\times$112$\times$5$\times$64\\
			pool$_1$&Max(1,2,2), stride (1,2,2)&56$\times$56$\times$5$\times$64\\
			\hline
			3d-conv$_2$&3$\times$3$\times$3$\times$128, padding(1,1,1), relu&56$\times$56$\times$5$\times$128\\
			pool$_2$&Max(2,2,2), stride (2,2,2)&28$\times$28$\times$2$\times$128\\
			\hline
			3d-conv$_{3a}$&3$\times$3$\times$3$\times$256, padding(1,1,1), relu&28$\times$28$\times$2$\times$256\\
			3d-conv$_{3b}$&3$\times$3$\times$3$\times$256, padding(1,1,1), relu&28$\times$28$\times$2$\times$256\\
			pool$_3$&Max(2,2,2), stride (2,2,2)&14$\times$14$\times$1$\times$256\\
			\hline
			3d-conv$_{4a}$&3$\times$3$\times$3$\times$512, padding(1,1,1), relu&14$\times$14$\times$1$\times$256\\
			3d-conv$_{4b}$&3$\times$3$\times$3$\times$512, padding(1,1,1), relu&14$\times$14$\times$1$\times$256\\
			pool$_4$&Max(1,2,2), stride (1,2,2)&7$\times$7$\times$1$\times$512\\
			\hline
			3d-conv$_{5a}$&3$\times$3$\times$3$\times$512, padding(1,1,1), relu&7$\times$7$\times$1$\times$512\\
			3d-conv$_{5b}$&3$\times$3$\times$3$\times$512, padding(1,1,1), relu&7$\times$7$\times$1$\times$512\\
			pool$_5$&Max(1,2,2), str(1,2,2), pad(1,0,0)&4$\times$4$\times$1$\times$512\\
			\hline
			fc$_1$&(8,192, 4,096)&1$\times$1$\times$4,096\\
			\hline
			fc$_2$&(4,096, 4,096)&1$\times$1$\times$4,096\\
			\hline\hline
			&\textbf{Tactile layers}&\textbf{Output size}\\
			\hline
			3d-conv$_6$&3$\times$3$\times$3$\times$8, padding(1,1,1), relu&4$\times$4$\times$10$\times$8\\
			pool$_6$&Max(2,2,2), stride (2,2,2)&2$\times$2$\times$5$\times$8\\
			\hline
			3d-conv$_7$&3$\times$3$\times$3$\times$16, padding(1,1,1), relu&2$\times$2$\times$5$\times$16\\
			pool$_7$&Max(2,2,2), stride (2,2,2)&1$\times$1$\times$2$\times$16\\
			\hline
			3d-conv$_8$&3$\times$1$\times$1$\times$32, padding(1,0,0), relu&1$\times$1$\times$3$\times$32\\
			pool$_8$&Max(2,1,1), stride (2,1,1)&1$\times$1$\times$1$\times$32\\
			\hline\hline
				&\textbf{Classification layers}&\textbf{Output size}\\
			\hline
			fc$_3$&(4096+32,128)&1$\times$1$\times$128\\
			\hline
			fc$_4$&(128, 3) max&1$\times$1$\times$1\\
			\hline
		\end{tabular}
	\end{center}
\end{table}

In practice, we use five $\footnote{The sequence length was selected by comparison experiments.}$ $112\times112\times3$ $\footnote{We selected this size as the default visual modal input size by comparing the performance of the model with different image sizes as input.}$ visual images and ten $4\times4\times3$ tactile images as input and the detailed network parameters are shown in Table \ref{c3d-vtfn-para}. The visual features extraction module includes five C3D and two FC layers. The convolution kernel size and stride size of each convolutional layer are not exactly the same. The output size of the final visual C3D layer is $4\times4\times1\times512$. The features from the visual C3D layers are fed to two FC layers and transformed into a 4096-dimensional feature vector. Similarly, the tactile features extraction module are composed of three C3D layers and followed by two FC layers. The difference is that the output tactile feature vector is 32-dimensional. Finally, the feature vectors from the two modalities are concatenated together to output the final grasp state category through two classification FC layers. 

Specifically, we use Xavier initialization \cite{c21} to initiate network weights and cross-entropy \cite{c22} as the loss function. Adam optimizer \cite{c23} with $1e-07$ learning rate is adopted in the training process. The model is implemented on the PyTorch platform $\footnote{Source code for study replication is available at: \url{https://github.com/swchui/Grasping-state-assessment}}$ and trained on an NVIDIA DGX server. The batch size is set as 8 in this paper.

\section{Experiments}
In this section, we first introduce our grasp state assessment dataset (GSA dataset) and the experimental setup. Then the performance comparison of C3D-VTFN with different structures and parameters on the GSA Dataset is provided.  Finally, we perform two delicate grasp experiments of a deformable object based on the C3D-VTFN model.
\subsection{The GSA dataset introduction}
All of the experiments are conducted with a 6-DOFs UR3 robot arm equipped with an OnRobot RG2 gripper. Specifically, one finger of the gripper is covered by a XELA tactile sensor and a 1080P USB camera is mounted on the top of the gripper as a wrist-camera. The robot setup is shown in Fig. \ref{robot_setup}. 

\begin{figure}[t!]
	\centering
	\includegraphics[width=8.3cm]{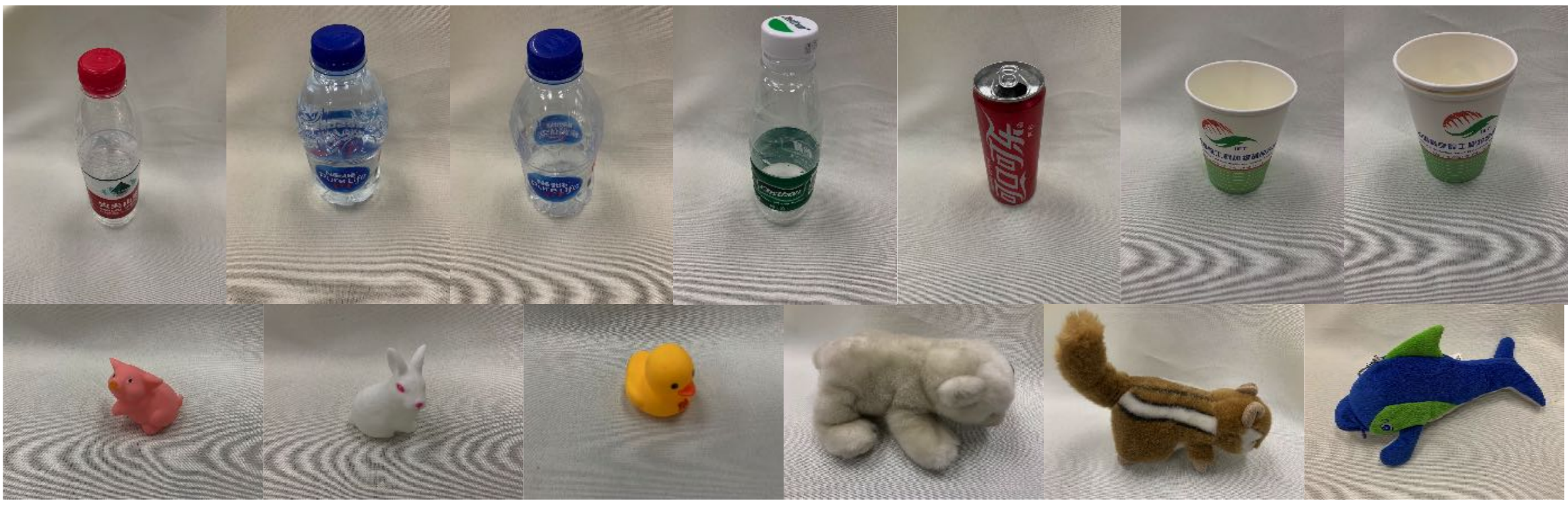}
	\caption{Some deformable objects of the GSA dataset. }
	\label{objects}
\end{figure}

The GSA dataset is built by extensive grasping and lifting experiments on 16 deformable objects of different sizes, shapes, textures, materials, and weights, some of them are shown in Fig. \ref{objects}. Inspired by \cite{c4}, different grasp widths and forces are selected to balance the number of routines with different labels. In this way, the grasp states are automatically labeled in each grasping and lifting trial. In each grasp experiment, an object is grasped with the preset width and force and lifted slowly for 20.0 mm (The lifting speed is set to 10.0 mm/s). During the grasping and lifting process, the data are collected by the visual sensor with a 30 Hz and tactile sensor with 60 Hz, respectively. We perform 50 to 60 grasps per object, collecting approximately 30 to 40 frames of visual images and 60 to 80 frames of tactile images per grasp trial. As a result, the GSA dataset consists of approximately 20,000 5-frame visual image sequences and corresponding tactile image sequence samples. Among them, the grasping data of randomly selected thirteen objects is used to train the model, and the grasping data of the remaining three objects is used for testing. The detailed GSA dataset is available at \url{https://github.com/swchui/Grasping-state-assessment/graspingdata}.
\subsection{Performance comparison results}
To evaluate the performance of the proposed model more comprehensively and accurately, we compared the Precision, Recall, F1 score, and model size of the model with different inputs and structures. The Precision, Recall, and F1 score are used to evaluate the classification performance, and the model size is adopted to compute the real-time performance of different models. The performance of the model may be affected by different length of input sequence, image size, inter-frame interval, and input of single-mode or dual-mode perception.
\subsubsection{Different input lengths} Longer sequences not only mean more temporal information, but also result in redundant calculations and bring more noise information. The visual sequence lengths of 3, 4, 5, 6, 7, and 8 are selected as the inputs of the model for comparative evaluation, and the tactile sequence length is set to twice the visual according to the reading frequency. The experiments results are shown in Table \ref{results_length}.
\begin{table}[thpb]
	\caption{Experimental results of the models with different input length.}
	\label{results_length}
	\begin{center}
		\begin{tabular}{|c|c|c|c|c|c|c|}
			\hline
			\multirow{2}*{ }&\multicolumn{6}{c|}{Sequence length}\\
			\cline{2-7}
			&3&4&\textbf{5}&6&7&8\\
			\hline
			Precision&75.78&95.21&\textbf{99.97}&99.80&90.27&90.13\\
			\hline
			Recall&67.27&95.49&\textbf{99.98}&99.74&88.81&79.91\\
			\hline
			F1 score&67.42&95.08&\textbf{99.98}&99.77&88.79&79.83\\
			\hline
			Size (M)&78.53&78.53&78.53&78.54&78.54&78.54\\
			\hline
		\end{tabular}
	\end{center}
\end{table}

The results show that it is not the more extended the input sequence is, the better the classification performance is. The model with a sequence length of 8 has a classification accuracy of 10\% lower than that of the sequence length of 5. As a result, the optimal classification performance is obtained when the sequence length is 5.

\subsubsection{Different inter-frame interval} Since sequences with different time intervals have different characteristics, the large inter-frame interval can result in a reduction in sample rate, and whether this affects the performance of the model is still a question worth exploring. Due to the reading speed of sensors is fixed, we set a \emph{basic} input sample in which the visual and tactile images are consecutive recorded reading. We also build \emph{step} input samples in which the data reading is selected with step 2 and 3. We set the other parameters as default, and only change the inter-frame interval for a comparison test. The results are shown in Table \ref{results_log}.

\begin{table}[thpb]
	\caption{Experimental results of the models with different input inter-frame intervals.}
	\label{results_log}
	\begin{center}
		\begin{tabular}{|c|c|c|c|}
			\hline
			\multirow{2}*{ }&\multicolumn{3}{c|}{Frame interval}\\
			\cline{2-4}
			&\textbf{1}&2&3\\
			\hline
			Precision&\textbf{99.97}&98.03&85.62\\
			\hline
			Recall&\textbf{99.98}&98.50&85.50\\
			\hline
			F1 score&\textbf{99.98}&98.23&83.80\\
			\hline
			Size (M)&78.53&78.53&78.53\\
			\hline
		\end{tabular}
	\end{center}
\end{table}

\begin{figure}[thpb]
	\centering
	\subfigure[Basic]{
		\label{log_image_a}
		\includegraphics[width=2.5cm]{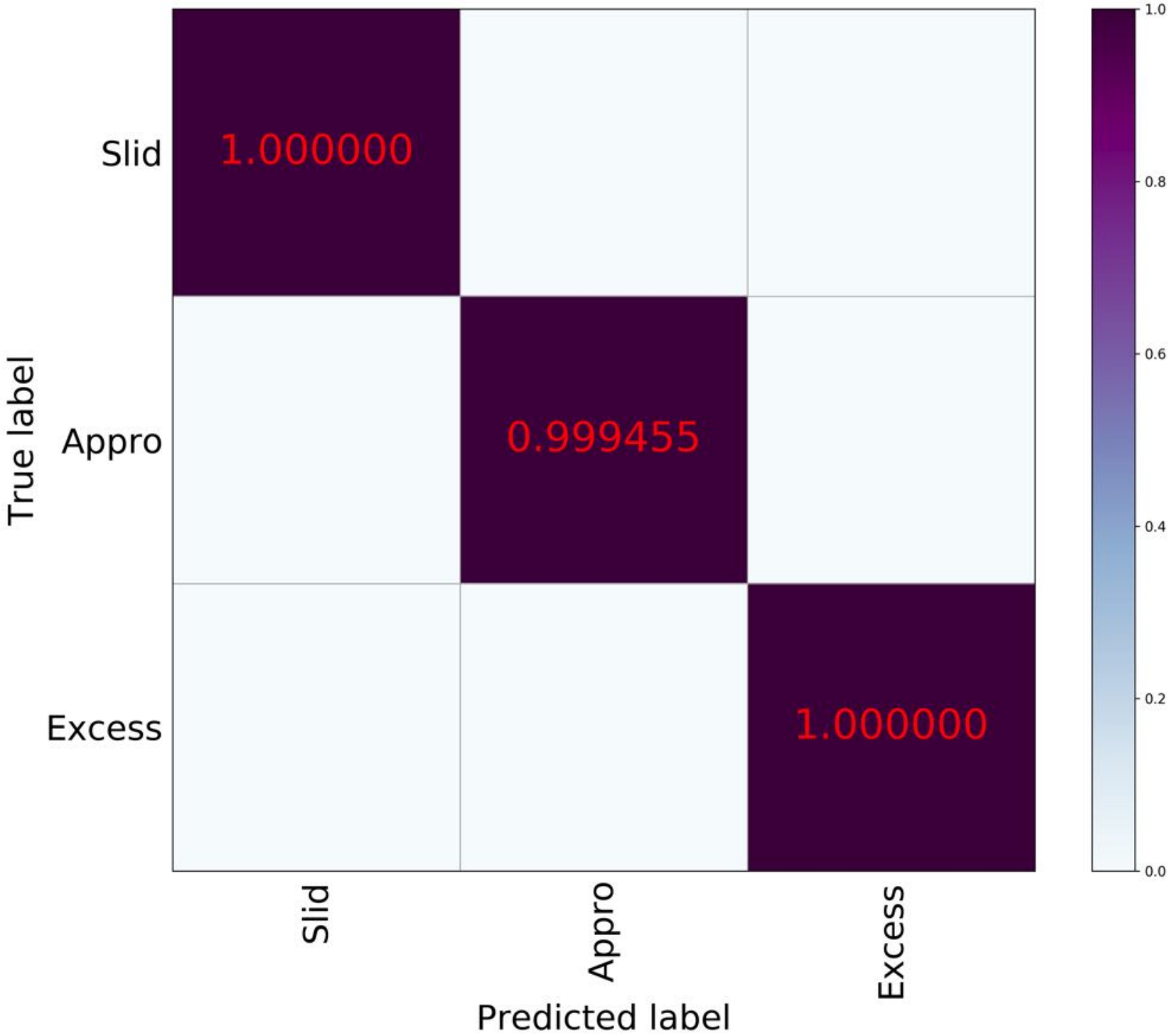}}
		\subfigure[Step 2]{
		\label{log_image_b}
		\includegraphics[width=2.5cm]{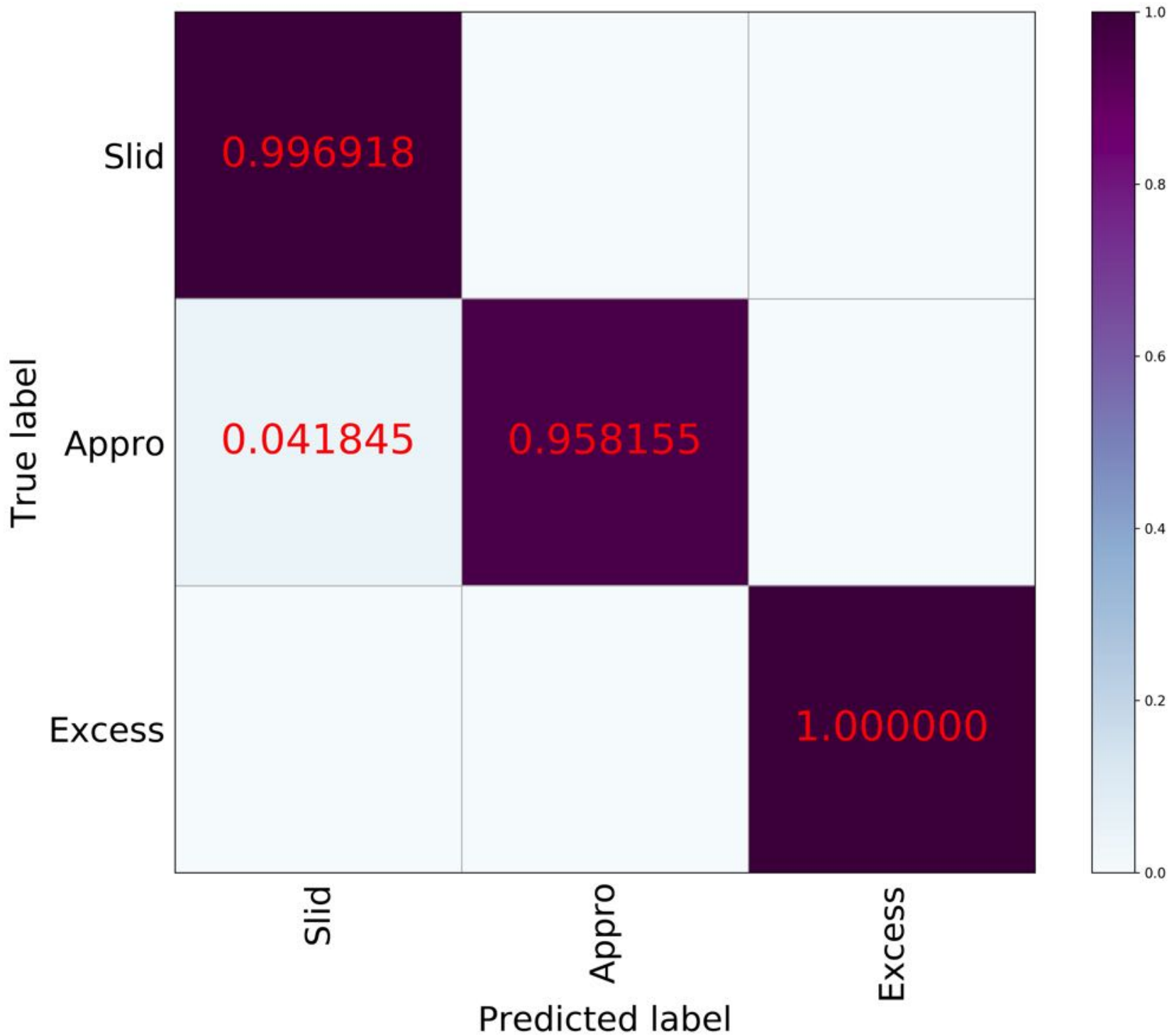}}
		\subfigure[Step 3]{
		\label{log_image_c}
		\includegraphics[width=2.5cm]{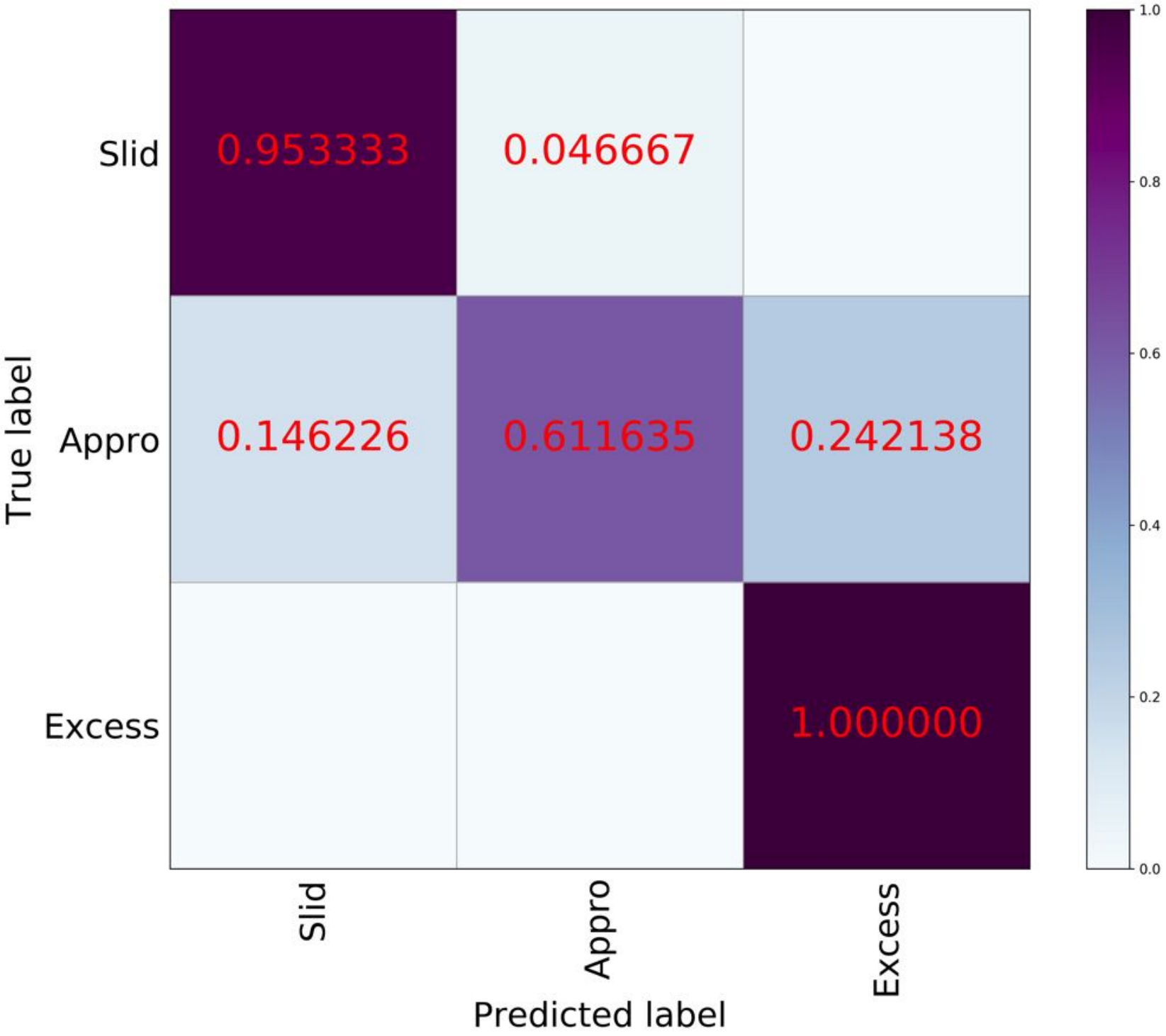}}
	\caption{The confusion matrices of models with different inter-frame interval.}
	\label{log_image}
\end{figure}

Table \ref{results_log} suggests that the using \emph{step} sampling method would be worse, especially for the appropriate grasping state. The confusion matrix shows that the grasp state of the appropriate is higher in the case where the step setting is smaller, as shown in Fig. \ref{log_image}. The intuitive explanation is that the reduction of the sample rate will reduce the confidence of the proposed model in determining the proper grasp state, which makes it more biased toward sliding or excessive state.
\subsubsection{Different image size} The image size directly determines the amount of information input by the visual modality and the amount of model parameters. Therefore, we set the inter-frame interval to 1, set  the input image sequence length to 5 (the tactile sequence length corresponds to 10), set the input image size to $32\times 32$, $64\times 64$, $112\times112$, $224\times224$, and $512\times 512$ respectively, and modify the corresponding model parameters, the results are shown in Table \ref{results_image}.
\begin{table}[thpb]
	\caption{Experimental results of the models with different image size.}
	\label{results_image}
	\begin{center}
		\begin{tabular}{|c|c|c|c|c|c|}
			\hline
			\multirow{2}*{ }&\multicolumn{5}{c|}{Image size}\\
			\cline{2-6}
			&32&64&\textbf{112}&224&512\\
			\hline
			Precision&76.78&90.01&\textbf{99.97}&89.36&73.44\\
			\hline
			Recall&66.68&80.05&\textbf{99.98}&89.12&69.86\\
			\hline
			F1 score&66.48&80.44&\textbf{99.98}&87.21&62.23\\
			\hline
			Size (M)&\textbf{55.53}&64.37&78.53&92.69&106.84\\
			\hline
		\end{tabular}
	\end{center}
\end{table}

\begin{figure}[thpb]
	\centering
	\subfigure[image size (32)]{
		\label{image_size_a}
		\includegraphics[width=2.5cm]{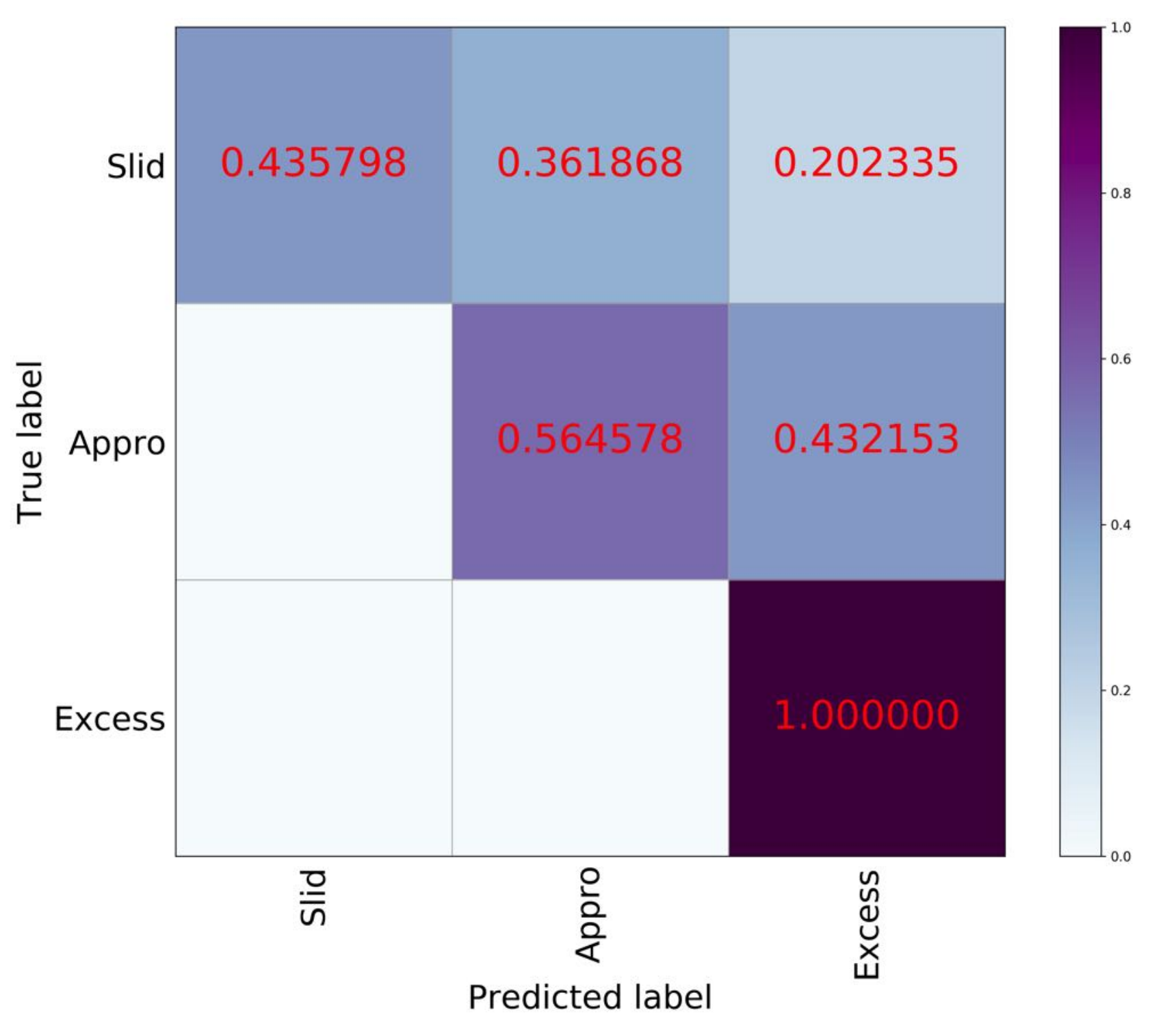}}
	\subfigure[image size (64)]{
		\label{image_size_b}
		\includegraphics[width=2.5cm]{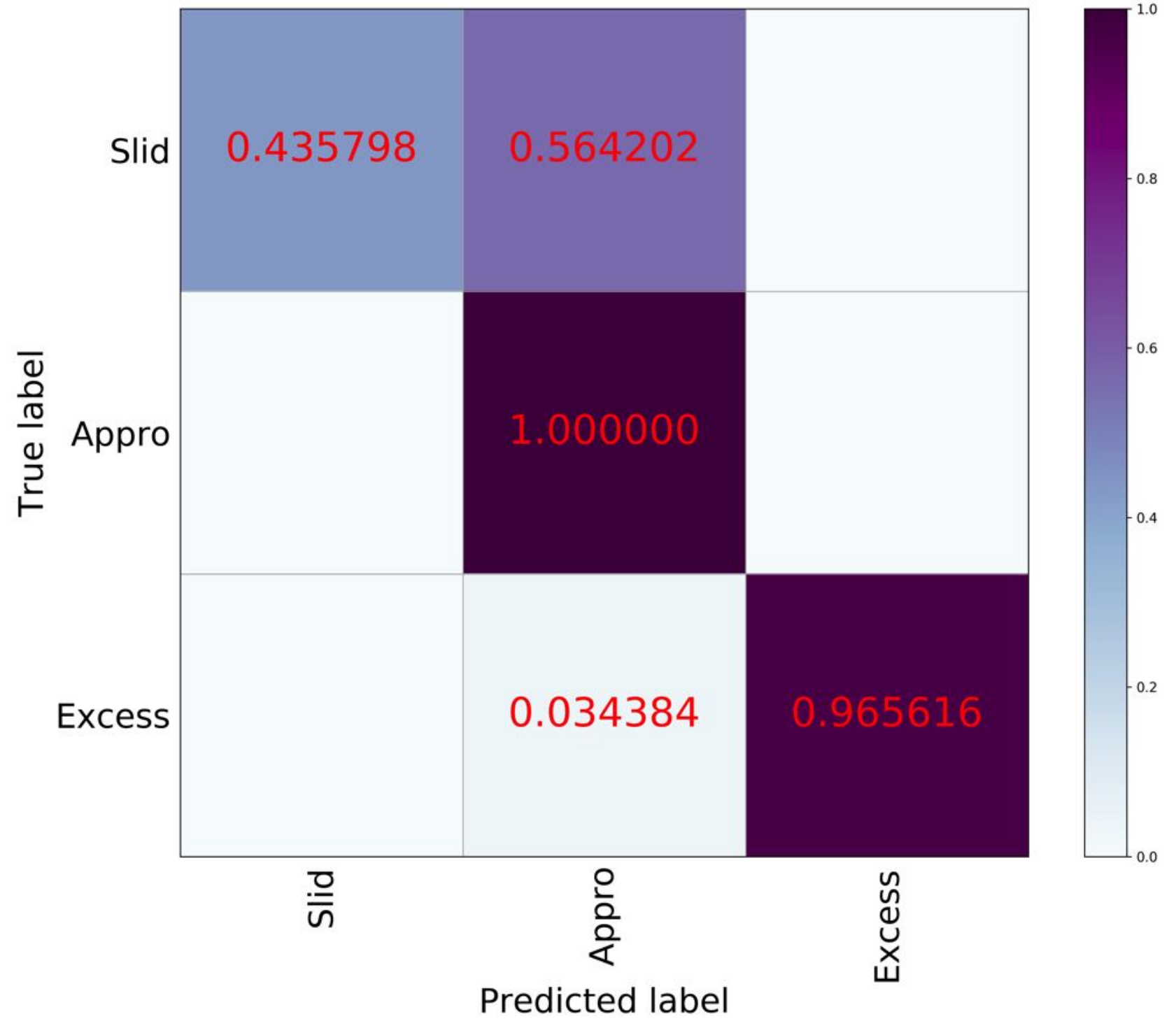}}
	\subfigure[image size (112)]{
		\label{image_size_c}
		\includegraphics[width=2.5cm]{figures/confusion_matrix_log_1.pdf}}
	\subfigure[image size (224)]{
		\label{image_size_d}
		\includegraphics[width=2.5cm]{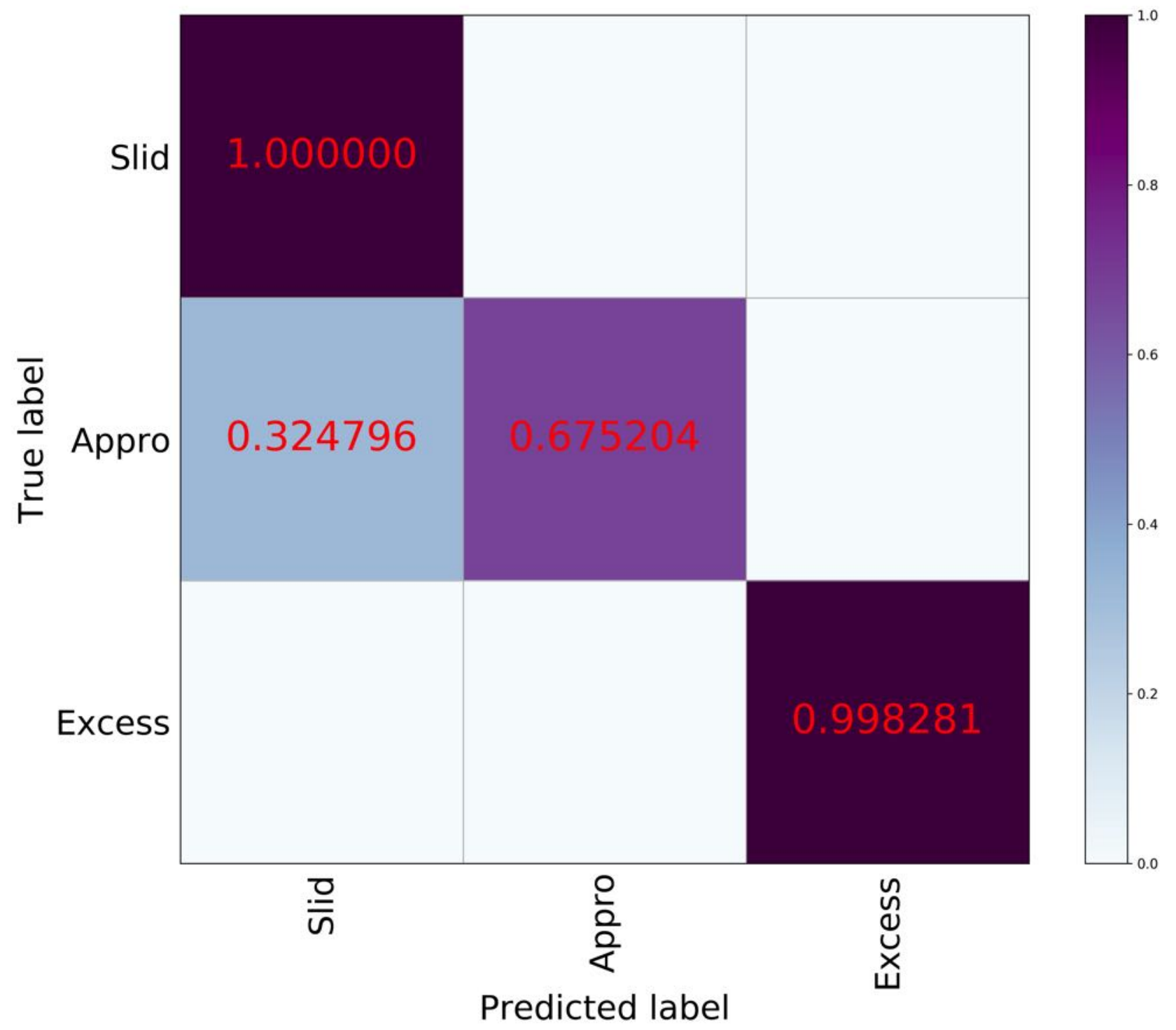}}
	\subfigure[image size (512)]{
		\label{image_size_e}
		\includegraphics[width=2.5cm]{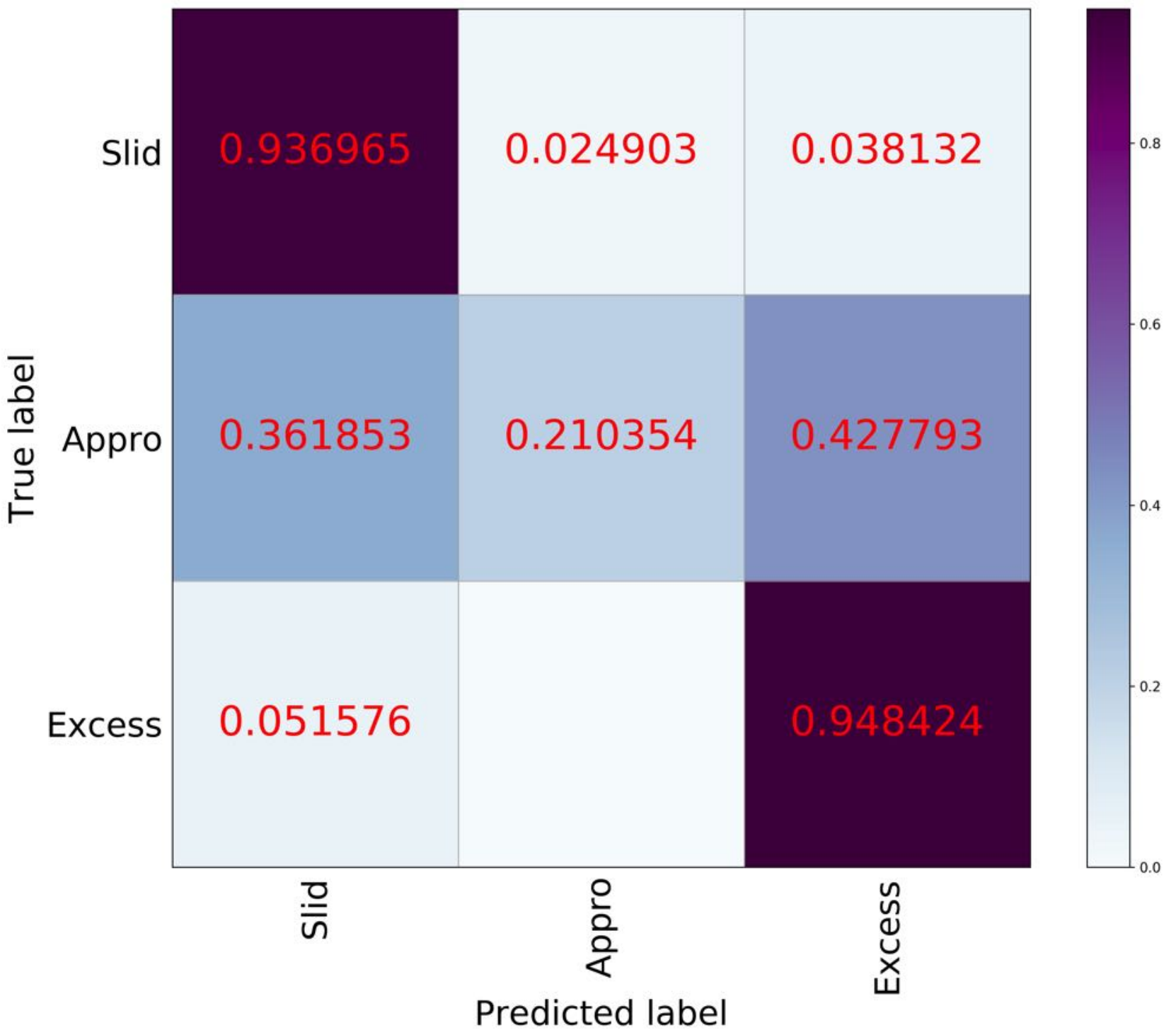}}
	\caption{The confusion matrices of models with different image size.}
	\label{image_size}
\end{figure}

The experimental results indicate that the model performance is not directly proportional to image size. We find that as the size of the image increases, the model detects the sliding state more accurately, but the appropriate grasp state detection performance becomes worse, as shown in Fig. \ref{image_size}. Hence, we select the visual sequence with image size 112 as the input to the model.

\subsubsection{Single-or-dual modal perception} Furthermore, different modal combinations are tested to verify the performance benefits of visual-tactile fusion perception. For the visual-only mode, we select the visual features extraction module and classification module from the C3D-VTFN model (as shown in Fig. \ref{modelarch}). Similarly, the tactile-only mode is tested by combining the tactile features extraction module and a classification module. The comparison results are shown in Table \ref{results_vtf}.
\begin{table}[thpb]
	\caption{Experimental results of the models with single-or-dual modal perception.}
	\label{results_vtf}
	\begin{center}
		\begin{tabular}{|c|c|c|c|}
			\hline
			&Tactile-only&Visual-only&Visual-Tactile fusion\\
			\hline
			Precision&70.30&79.74&\textbf{99.97}\\
			\hline
			Recall&72.36&79.77&\textbf{99.98}\\
			\hline
			F1 score&67.11&79.27&\textbf{99.98}\\
			\hline
			Size (M)&\textbf{0.01}&78.52&78.53\\
			\hline
		\end{tabular}
	\end{center}
\end{table}

\begin{figure}[thpb]
	\centering
	\subfigure[Visual-only]{
		\label{visual_only_a}
		\includegraphics[width=2.5cm]{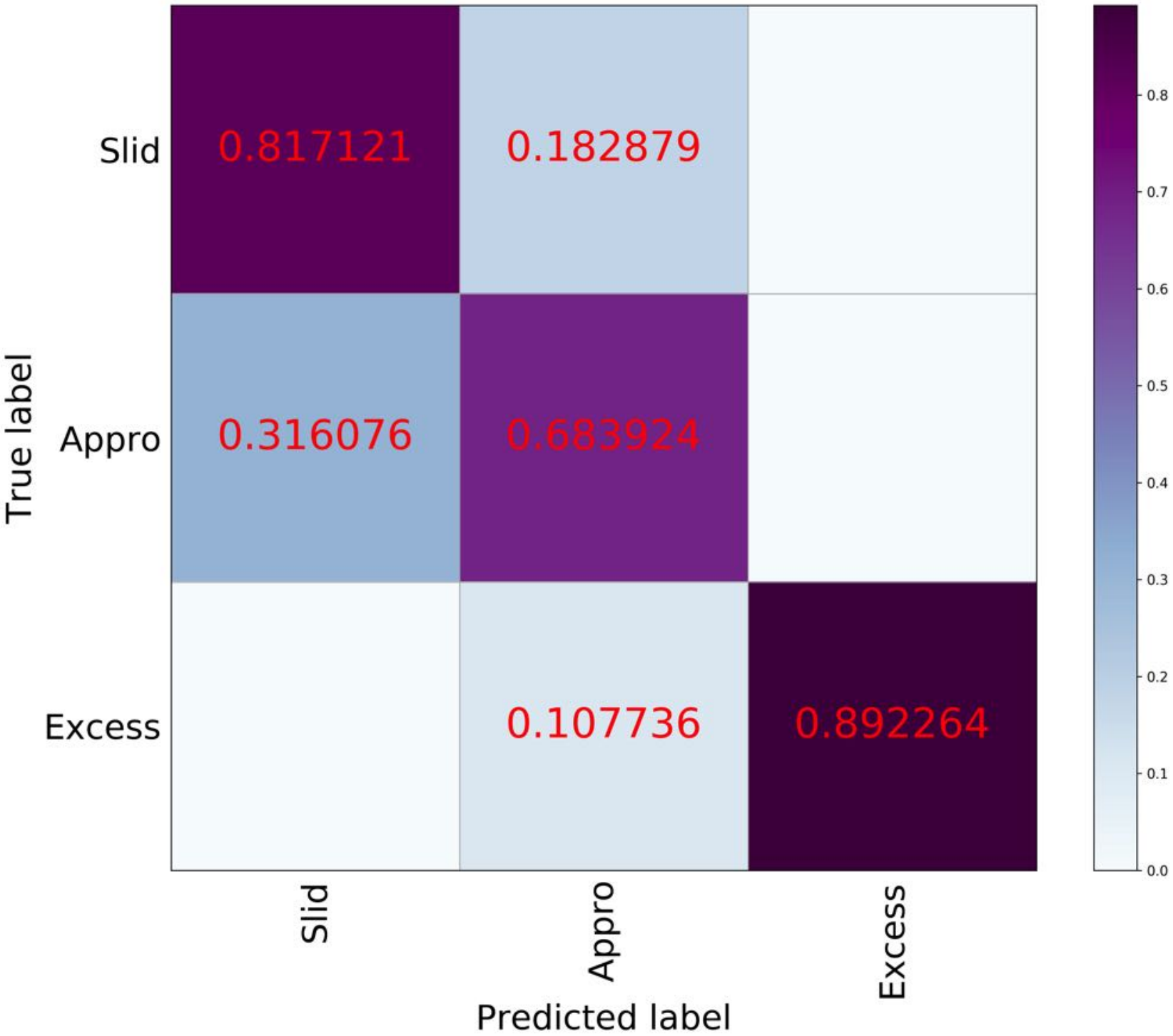}}
	\subfigure[Tactile-only]{
		\label{tactile_only_b}
		\includegraphics[width=2.5cm]{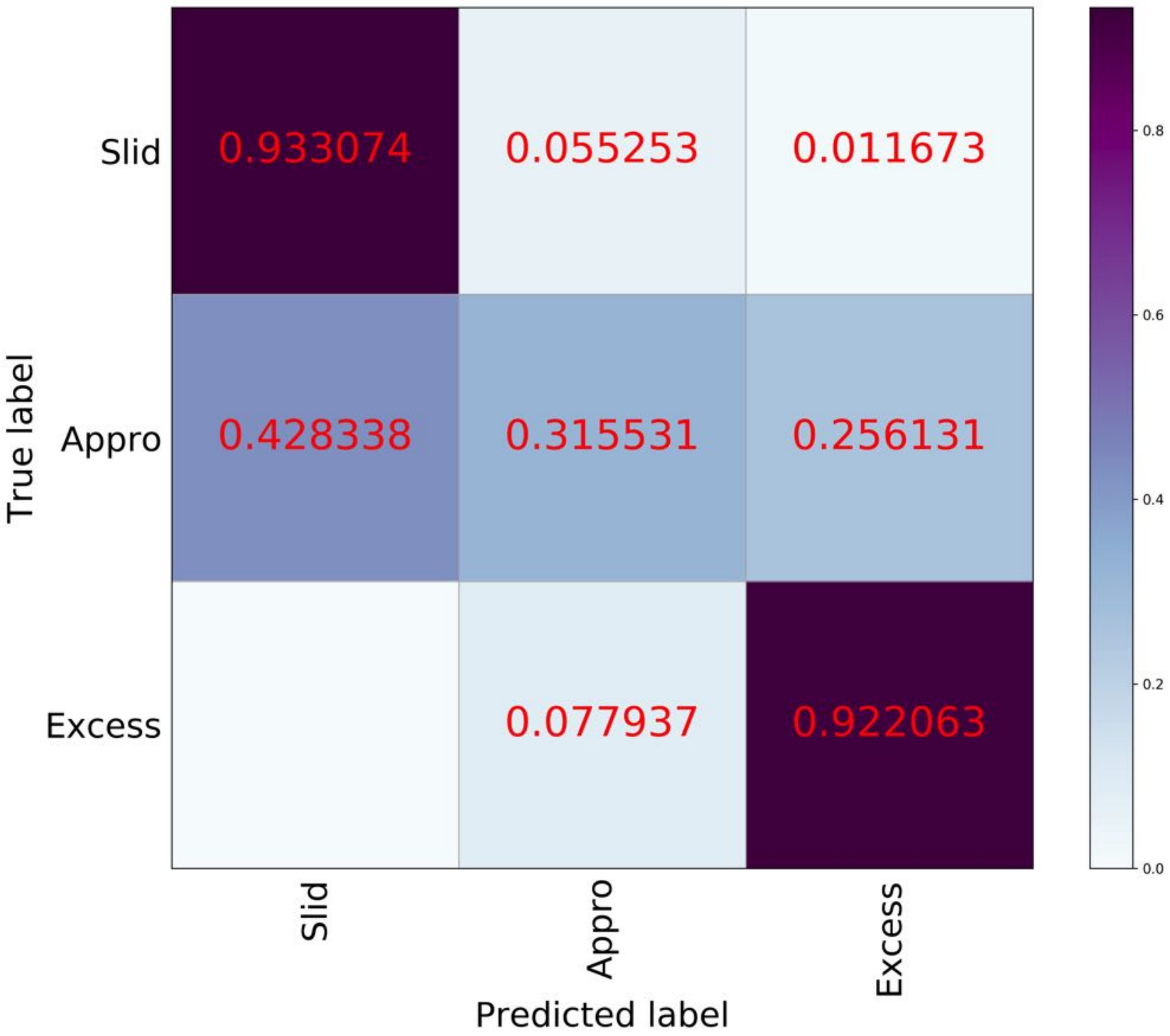}}
	\subfigure[VTF]{
		\label{vtf_c}
		\includegraphics[width=2.5cm]{figures/confusion_matrix_log_1.pdf}}
	\caption{The confusion matrices of models with different inputs.}
	\label{vtf_con_mat}
\end{figure}

According to the experimental results, visual-tactile fusion perception achieves much better precision, recall, and F1 score than that of any single modal perception. Theoretically, the visual image provides the geometrical information of the contact situation, which would have a better ability to distinguish the excessive grasp state from others. This analysis has been verified by the confusion matrices shown in Fig. \ref{visual_only_a} and Fig. \ref{tactile_only_b}. Meanwhile, the confusion matrices also show that the tactile-only model achieves better detection performance of the sliding grasp state than the visual-only model.

\subsection{Delicate grasp experiments based on C3D-VTFN model.}
Two delicate grasp experiments in this section are performed to further verify the effectiveness of the proposed model. We have developed a roughly grasp adjustment strategy that adjusts grasp width and force in real-time based on the grasp state detector (C3D-VTFN).  The detailed grasp regulation strategies are as follows,
$$ w_{t+1},f_{t+1}=\left\{
\begin{array}{rcl}
w_t-1,f_t+1       &      & {c_t=sliding,}\\
w_t,f_t    &      & {c_t=appropriate,}\\
w_t+1,f_t+1     &      & {c_t=excessive.}
\end{array} \right. $$

Where $w_t$ and $f_t$ represent the grasp width and force at the current moment, and $w_{t+1}$ and $f_{t+1}$ represent the next moment. Also, the adjustment of the grasp settings depends on the evaluation of the grasp state of the current moment $c_t$. 

On the one hand, a grasp adjustment experiment that begins with a slip is first performed, as described in Section \ref{sliding_grasp}. On the other hand, a grasp adjustment experiment with an initial excessive-force grip is performed, as described in Section \ref{excessive_grasp}.
\subsubsection{The sliding grasp experiments}
\label{sliding_grasp}
\begin{figure}[thpb]
	\centering
	\includegraphics[width=7.5cm]{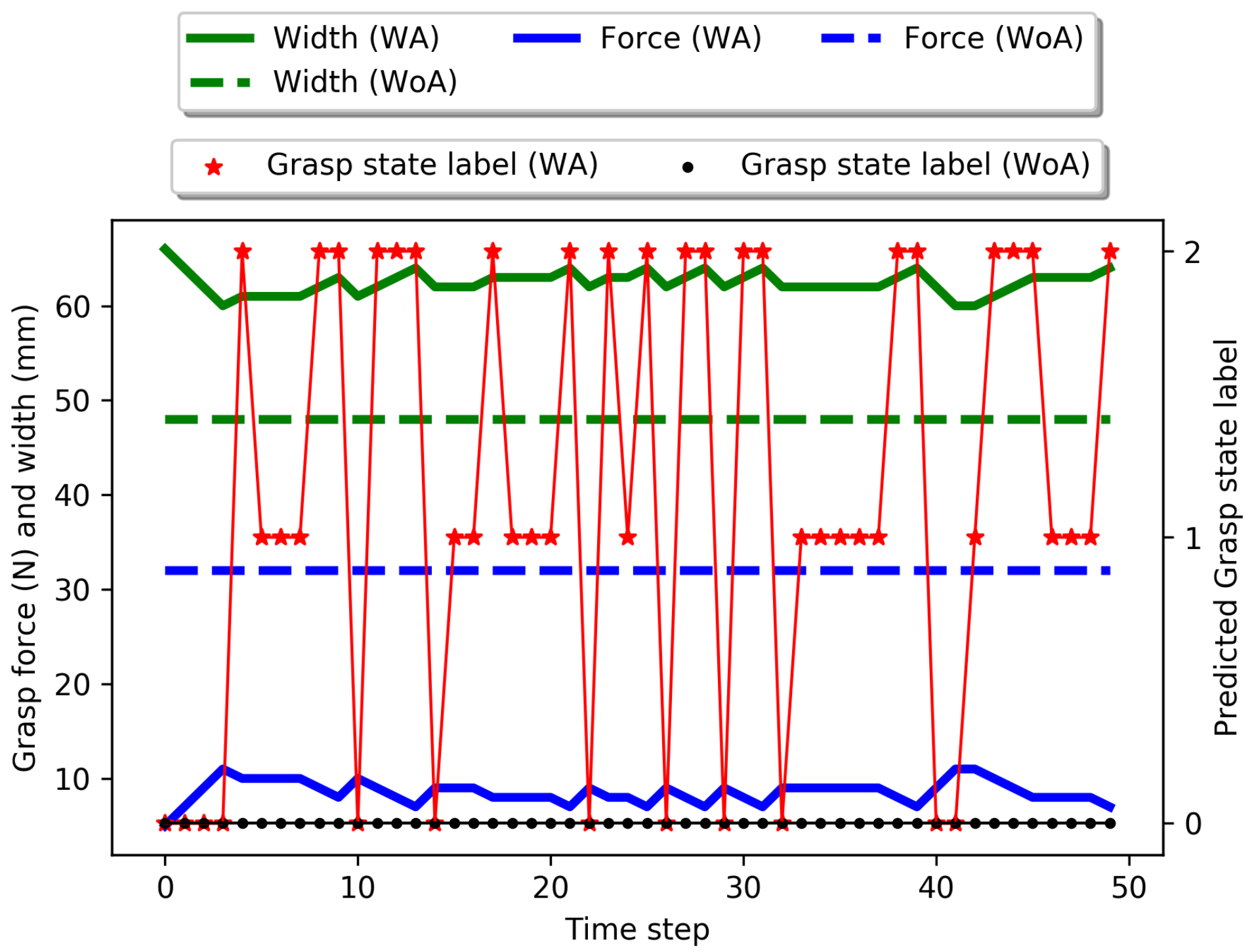}
	\caption{The real-time changing curve of different values in the two sliding grasp experiments. Solid line: the grasp process with adjustment. Dashed line: the grasp process without adjustment. The second axis labels: (0) sliding, (1) appropriate, (2) excessive.}
	\label{slip_ex_0}
\end{figure}

First, we set the grasp force to 5N and the grasp width to 66 mm for the grasping and lifting experiments of a deformable bottle (not included in the GSA dataset) with (WA) and without (WoA) adjustment strategy. In these two experiments,  the real-time changing curve of the adjustment grasp force, grasp width, and the detection grasp state is shown in Fig. \ref{slip_ex_0}.

Fig. \ref{slip_ex_0} shows that the proposed model can accurately detect the current state of the grasp state regardless of there is a grasp strategy adjustment. In the experiment without the adjustment strategy, the detector always detected the grasp state as \emph{sliding (0)}. However, in the experiment with the adjustment strategy, the detection state changes with the change of the grasp settings. In the grasp process with adjustment, the model still detects it as a sliding state even if has been adjusted at the beginning. The reason is that the preset grip width (68mm) is larger than the actual diameter of the bottle and takes a few steps to adjust. After a few steps, the entire lifting process is adjusted between the three grasp states according to the actual grasp situation to achieve a delicate grasp.
\subsubsection{The excessive grasp experiments}
\label{excessive_grasp}
\begin{figure}[thpb]
	\centering
	\includegraphics[width=7.5cm]{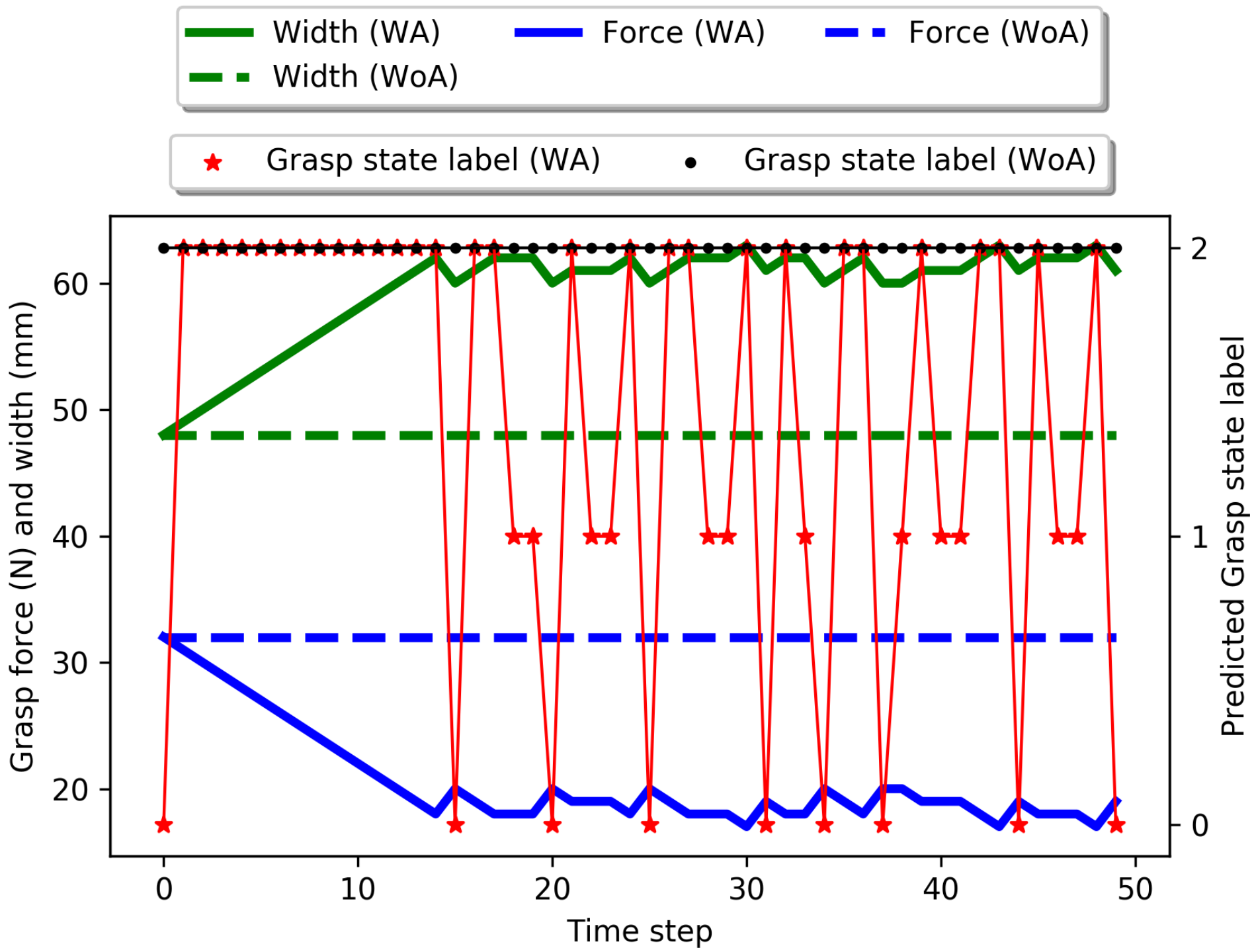}
	\caption{The real-time changing curve of different values in the two excessive grasp experiments.}
	\label{excessive_ex_0}
\end{figure}

Similar to the previous experiments, we preset the grasp force and width to 32N and 48mm, respectively, in the excessive grasp experiments. The changing curves of different grasp values are shown in Fig. \ref{excessive_ex_0}.

The grasp process in the experiment with adjustments can finally stabilize the grasp force and width, similar to those in the sliding experiment. Since the grasp force and width adjustment settings are very rough in this paper, the final steady-state of the two experiments is not wholly consistent. But it is sufficient to verify the performance of the proposed model. Please note that the retardation of the predicted grasp state at the beginning of the grasp adjustment experiment is due to the experimental settings and the initial grasp force setting being too large. 

The above two experiments have well verified the evaluation performance of the proposed model on the current grasp state. However, this adjustment strategy is not enough for delicate grasp, and it is necessary to set a more fine-grained adjustment strategy (See supplementary video materials for more experimental details).

Additionally, due to the fixed angle of view during data collection, the trained model has higher accuracy at fixed angles and less than ideal performance at other views. A feasible way is adding multiple different views of grasp settings in each experiment, which significantly increases the scale of the GSA dataset, and makes the model more generalized. Fortunately, this limitation does not prevent us for verifying the feasibility of the C3D-based visual-tactile fusion perception approach.

\section{Conclusion and future work}
A network named C3D-VTFN is proposed to assess the grasp state of various deformable objects by using visual-tactile fusion perception in this paper. We extract the features of the visual and tactile modalities by 3D convolution layers, which provides a new feature extraction scheme for the visual-tactile fusion perception tasks. Besides, the GSA dataset used to train and test the proposed model is established by extensive grasping and lifting experiments in this paper, and the experimental results show the effectiveness and high accuracy of the proposed model. Finally, we perform two delicate grasp experiments with a rough adjustment strategy based on the proposed model and achieved convincing results.

In the future, we will explore perceptual models that are more in line with human vision-tactile fusion properties and their applications in robotic grasping and manipulation.



%

%


\newpage
\balance

\balance
\end{document}